%% file: main.tex
\documentclass[conference]{IEEEtran}
\IEEEoverridecommandlockouts

\input{setup/package}
\input{setup/macros}

\input{setup/symbols}

\input{setup/graphicspath}
\usepackage{cite}
\usepackage{amsmath,amssymb,amsfonts}
\usepackage{algorithmic}
\usepackage{graphicx}
\usepackage{textcomp}
\usepackage{xcolor}
\newcommand{\mname}{SMILES-Mamba}
\newcommand{\std}[1]{\scriptsize{$\pm$#1}}

\def\BibTeX{{\rm B\kern-.05em{\sc i\kern-.025em b}\kern-.08em
    T\kern-.1667em\lower.7ex\hbox{E}\kern-.125emX}}
\begin{document}

\title{\mname: Chemical Mamba Models for Drug ADMET Prediction}







\title{\mname: Chemical Mamba Foundation Models for Drug ADMET Prediction}

\author{Bohao Xu$^{1,*}$, Yingzhou Lu$^{2,*}$, Chenhao Li$^{3}$, Ling Yue$^{1}$, Xiao Wang$^{4}$, Tianfan Fu$^{1}$, Minjie Shen$^{5}$, Lulu Chen$^{5}$ \\ 
1. Rensselaer Polytechnic Institute, 2. Stanford University, 3. University of Illinois Urbana-Champaign\\  4. University of Washington, 
5. Virginia Tech }


\maketitle

\begin{abstract} 
In drug discovery, predicting the absorption, distribution, metabolism, excretion, and toxicity (ADMET) properties of small-molecule drugs is critical for ensuring safety and efficacy. However, the process of accurately predicting these properties is often resource-intensive and requires extensive experimental data. To address this challenge, we propose \mname, a two-stage model that leverages both unlabeled and labeled data through a combination of self-supervised pretraining and fine-tuning strategies. The model first pre-trains on a large corpus of unlabeled SMILES strings to capture the underlying chemical structure and relationships, before being fine-tuned on smaller, labeled datasets specific to ADMET tasks. Our results demonstrate that \mname~exhibits competitive performance across 22 ADMET datasets, achieving the highest score in 14 tasks, highlighting the potential of self-supervised learning in improving molecular property prediction. This approach not only enhances prediction accuracy but also reduces the dependence on large, labeled datasets, offering a promising direction for future research in drug discovery. The code and pretrained model would be released after acceptance. 
\end{abstract}

\begin{IEEEkeywords}
Drug ADMET Prediction, Drug Discovery, Foundation Model, Pretraining
\end{IEEEkeywords}

\section{Introduction}
Small-molecule drugs are chemical compounds with desirable pharmaceutical properties. 
After being taken orally, it needs to travel from the site of administration (e.g., oral) to the site of action (e.g., a tissue), then decomposes and is finally excreted from the body~\cite{hou2007adme,dong2018admetlab}. 
To do that safely and efficaciously, the chemical is required to have numerous ideal absorption, distribution, metabolism, excretion, and toxicity (ADMET) properties. 
Small-molecule ADMET (including absorption, distribution, metabolism, excretion, and toxicity) properties are crucial to drugs' safety in the human body. A poor ADMET profile is the major reason for failure in pre-clinical and early clinical trial phases~\cite{yue2024ct,yue2024trialdura,yue2024trialenroll,chen2024trialbench,wang2024twin}. Early and accurate ADMET characterization is necessary for the successful development of small-molecule drug candidates during the drug discovery stage~\cite{fu2022hint,fu2023automated}.

In recent years, machine learning models have become increasingly important in predicting molecular properties, offering a way to prioritize potentially desirable molecules without the need for extensive and resource-intensive wet-lab experiments~\cite{huang2020deeppurpose,huang2021therapeutics}. This approach can significantly accelerate the drug discovery process, saving time and resources while improving the chances of identifying viable drug candidates. However, traditional models often struggle with the complexity and variability inherent in ADMET prediction, necessitating the development of more sophisticated approaches.

This paper introduces \mname, a two-stage model designed to enhance molecular property prediction by leveraging both unlabeled and labeled data through self-supervised learning-based pretraining followed by fine-tuning. By learning from a vast corpus of unlabeled molecular data, such as SMILES strings, during the pretraining stage, \mname~captures underlying chemical structures and relationships, which are then fine-tuned on specific ADMET tasks using labeled datasets. Our results demonstrate that \mname~outperforms several state-of-the-art methods across a range of ADMET datasets, highlighting the potential of self-supervised learning in advancing molecular property prediction and providing a promising direction for future research in drug discovery.

Our contributions of this paper could be summarized as:
\begin{itemize}[leftmargin=*]
\item We propose a two-stage (pre-training and fine-tuning) model \mname~to utilize both unlabeled data and labeled data to have better molecular properties prediction performance.
\item \mname~has better performance, outperforming a series of state-of-the-art methods on most of the ADMET datasets, obtaining the highest score in 14 tasks among all the 22 tasks. 
\end{itemize}

\section{Problem Statement}

\subsection{Drug Representation: SMILES String}
\label{sec:smiles}
\begin{figure}
 \centering
 \includegraphics[width=0.92\linewidth]{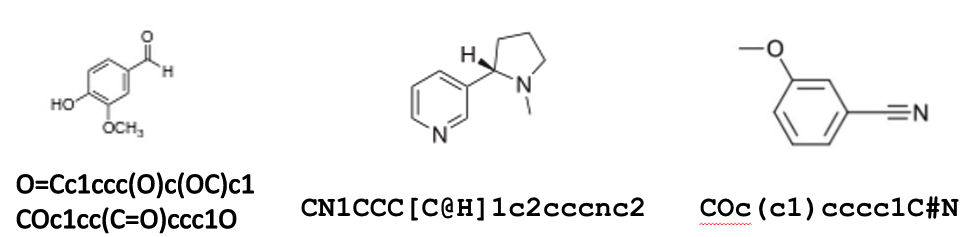}
 \caption{Some examples of SMILES strings. }
 \label{fig:smiles}
\end{figure}

The natural idea is to represent the chemical compound in a string of atoms, which is a convenient format for storage. Weininger et al.~\cite{weininger1988smiles} invented SMILES (Simplified Molecular Input Line Entry System) in the 1980s, which has later been optimized and extended. 
The simplified molecular-input line-entry system (SMILES) is a specification in the form of a line notation for describing the structure of chemical species using short ASCII strings. 
To date, the SMILES string has become the most standard representation of chemical molecules. 
We show some examples of SMILES in Figure~\ref{fig:smiles}. 

\subsection{Drug Pharmaceutical Property}
\label{sec:property}

In drug discovery, we need to assess a chemical compound on various pharmaceutical properties. For example, the properties evaluate whether the chemical compound is toxic to the human body, or whether the chemical compound can be absorbed by the human body. 

Among all the drug properties of interest, pharmacokinetics (PK) and pharmacodynamic (PD) properties are important ones that measure how a drug interacts with the body as a whole~\cite{ghosh2016modeling} and are keys to the safety of a drug. 
Pharmacokinetics focuses on the movement of drugs through the human body, whereas pharmacodynamics refers to the body's biological response to drugs. 
Evaluating drug molecules' PK/PD experimental scores requires intensive wet-lab experiments. 
The most useful PK/PD properties include the following (ADMET):
\begin{itemize}[leftmargin=*]
\item {\bf Absorption (A):} The absorption model describes how drugs are absorbed into the human body to reach the site of action~\cite{wu2022cosbin}. A poor-absorption drug is usually less desirable. 
\item {\bf Distribution (D):} The drug distribution model measures the ability of the molecule to move through the bloodstream to various parts of the body. A stronger distribution movement is desirable. 
\item {\bf Metabolism (M):} The drug metabolism rate determines the duration of a drug's efficacy. 
\item {\bf Excretion (E):} The drug excretion rate measures how efficiently a drug's toxic components can be removed from the body. 
\item {\bf Toxicity (T):} The drug toxicity measures the damage a drug can cause to the human body.
\end{itemize}

\subsection{Drug ADMET Property Prediction}

Predicting molecular properties with machine learning models can help us prioritize potentially desirable molecules without wet lab experiments, which would save a large number of resources. 
Thus, it is a fundamental task in drug discovery and is formulated as 
\begin{equation}
\label{eqn:property_pred}
{y} = f_{\mathbf{\theta}}(X), 
\end{equation}
where $X$ represents the drug molecule, $y$ denotes the predicting target, for the regression task, $y\in \mathbb{R}$ is the continuous value, while for the classification task, $y$ is a categorical label, e.g., $y\in\{0,1\}$ for binary classification; $f_{\mathbf{\theta}}$ is the machine learning model with learnable parameters $\mathbf{\theta}$, e.g., $f_{\mathbf{\theta}}$ can be \mname, graph neural network~\cite{kipf2016semi}, recurrent neural network~\cite{hochreiter1996lstm} or logistic regression~\cite{huang2020deeppurpose}; 
molecular property prediction can be used to help/accelerate the virtual screening process.

\section{Method: \mname}

\subsection{Overview} 
The \mname~model employs a two-stage approach consisting of pre-training and fine-tuning to enhance the prediction of molecular properties by effectively leveraging both unlabeled and labeled data. We first describe the basic Mamba model in Section~\ref{sec:mamba}. The pretraining and finetuning steps are described in Section~\ref{sec:pretrain} and Section~\ref{sec:finetune}, respectively.

\subsection{Model Backbone: Mamba}
\label{sec:mamba} 

Mamba~\cite{gu2023mamba} is a specialized implementation of the Structured State Space Sequence (S4) model designed for effectively handling long-range dependencies in sequential data. Unlike traditional models, Mamba excels in tasks like time-series analysis and natural language processing by capturing both local and global temporal patterns within sequences. It leverages state space models to maintain and update hidden states over extended sequences, ensuring accurate modeling of complex temporal dynamics. Mamba's architecture supports efficient parallel processing, making it scalable for large datasets, and is particularly useful in applications where understanding long-term dependencies is critical. Mamba architecture has been successfully applied in many practical applications, such as natural language processing~\cite{waleffe2024empirical,yue2024biomamba}, computer vision~\cite{yu2024mambaout,wang2024large}, EEG analysis~\cite{xu2024mambacapsule}. 

Transformers~\cite{vaswani2017attention} and Mamba are both powerful models for handling sequential data, but they differ significantly in their approaches and strengths. Transformers rely on self-attention mechanisms to capture dependencies within sequences, excelling at tasks like natural language processing and machine translation due to their ability to model relationships between all elements in a sequence simultaneously. However, Transformers can struggle with very long sequences due to their computational complexity. In contrast, Mamba, based on the Structured State Space Sequence (S4) model, is specifically designed to handle long-range dependencies efficiently by leveraging state space models that maintain and update hidden states over extended sequences. This makes Mamba particularly well-suited for tasks like time-series analysis, which is crucial for capturing long-term temporal patterns. While Transformers offer versatility and strong performance in a variety of tasks, Mamba dominate in scenarios where long-range dependencies are key and where computational efficiency over long sequences is required.

\subsection{Pretraining: Property-agnostic Mamba Model}
\label{sec:pretrain}
Pretraining is essential because it allows a model to learn general features and patterns from large datasets, which can then be fine-tuned for specific tasks with smaller labeled datasets. This process significantly improves the model's performance, reduces the amount of labeled data needed, and accelerates training for downstream tasks by starting from a well-initialized state rather than from scratch. 

In the pre-training stage, the model is trained on a vast corpus of unlabeled molecular data, such as SMILES  strings, to learn the underlying chemical structures and relationships. This stage allows the model to develop a rich representation of molecular features without the need for explicit labels, capturing essential patterns and dependencies in molecular data.
The Mamba model is an autoregressive model, and the pretraining objective is next-step prediction. The dataset does not contain any label about ADMET property, thus, the pretrained Mamba model is property-agnostic. 

We use ZINC~\cite{sterling2015zinc} (a well-known druglike small-molecule library) to pretrain the \mname~model. ZINC is a free database of commercially available compounds for virtual screening~\cite{sterling2015zinc}. 
It comprises over 230 million purchasable compounds in 3D formats. We use a 250K sampled version. The ZINC dataset does not contain any molecular properties. 
ZINC dataset can be used in 
\begin{enumerate}
\item {Pretraining property-agnostic Mamba model}. 
\item {Collecting basic vocabulary of tokens}. The token vocabulary includes ``C'', ``c'', ``O'', ``o'', ``N'', ``n'', ``S'', ``='', ``\#'', ``('', ``)'', ``['', ``]'', ``1'', ``2'', ``3'', etc.  
\end{enumerate}

\subsection{Fine-tuning: Property-specific Mamba Model}
\label{sec:finetune}

Once pre-trained, the model undergoes fine-tuning using a smaller, labeled dataset specific to the target task, such as predicting molecular properties like solubility, binding affinity, or toxicity. Fine-tuning adjusts the pre-trained model's parameters to optimize performance on the specific task, using the labeled data to refine and improve the model's predictions. This two-stage process significantly enhances the model's ability to predict molecular properties by combining the generalization capabilities learned during pre-training with the task-specific insights gained during fine-tuning.

By utilizing both unlabeled and labeled data, the \mname~model achieves superior prediction performance, making it a powerful tool in drug discovery and other applications requiring accurate molecular property predictions. This approach not only improves the efficiency of model training but also reduces the reliance on large amounts of labeled data, which can be scarce and costly to obtain.

\section{ADMET Datasets}
\label{sec:admet_dataset} 
We use 22 ADMET datasets. 
These datasets include metabolism with diverse types of CYP enzymes, half-life, clearance, and off-target effects. Real-world discovery studies drug candidates with diverse structures, and the ADMET benchmark datasets represent distribution shifts faced in the wild.
The detailed description of ADMET datasets are 
\begin{itemize}[leftmargin=*]
\item {\bf Absorption (A):} 
\begin{itemize}[leftmargin=*]
\item Caco2. The human colon epithelial cancer cell line, Caco2, is used as an in vitro model to simulate the human intestinal tissue. The experimental result on the rate of drug passing through the Caco-2 cells can approximate the rate at which the drug permeates the human intestinal tissue~\cite{wang2016adme}. It consists of 906 drug molecules with continuous labels. 
\item HIA. In the case of oral administration, a drug must be absorbed from a human's gastrointestinal system into the body's bloodstream. This process, known as human intestinal absorption (HIA), is critical in ensuring that the drug reaches its intended target. \cite{hou2007adme} collects an HIA database that comprises 578 drug molecules with binary labels. 
\item Pgp. P-glycoprotein (Pgp) is a protein belonging to the ATP-binding cassette (ABC) transporter family. It plays a significant role in processes such as intestinal absorption, drug metabolism, and brain penetration. Inhibition of Pgp can have profound effects on the bioavailability and safety of a drug~\cite{broccatelli2011novel}. The dataset contains 1,212 drug molecules with binary labels.
\item Bioav. Oral bioavailability is the term used to describe the speed and extent to which the active ingredient or active moiety of a drug product is absorbed from the gastrointestinal tract and becomes accessible at the intended site of action~\cite{ma2008prediction}. The dataset contains 640 drug molecules with binary labels.  
\item Lipo. Lipophilicity quantifies the capacity of a drug to dissolve in lipid-based environments, such as fats and oils. Drugs with high lipophilicity typically exhibit rapid metabolism, limited solubility, high turnover, and low absorption rates~\cite{wenlock2015experimental}. It contains 4,200 drug molecules with continuous labels. 
\item AqSol. Aqueous solubility refers to the capability of a drug to dissolve in water. Insufficient water solubility can result in slow drug absorption, inadequate bioavailability, and even potential toxicity. It is noteworthy that over 40\% of newly developed chemical entities exhibit poor solubility~\cite{sorkun2019aqsoldb}. The dataset consists of 9,982 drug molecules with continuous labels.   
\end{itemize}
\item {\bf Distribution (D):} 
\begin{itemize}[leftmargin=*]
\item BBB. The blood-brain barrier (BBB) serves as a protective barrier between the circulating blood and the brain's extracellular fluid, preventing the entry of most foreign drugs. Therefore, the ability of a drug to cross this barrier and reach the desired site of action presents a significant challenge in the development of drugs for the central nervous system~\cite{martins2012bayesian}. The dataset involves 1,975 drug molecules with binary labels. 
\item PPBR. The human plasma protein binding rate (PPBR) refers to the percentage of a drug that binds to plasma proteins in the blood. This rate significantly impacts the efficiency of drug delivery. When a drug has a lower binding rate, it can traverse and diffuse more effectively to reach its target site. This information is derived from a ChEMBL assay conducted by AstraZeneca~\cite{wenlock2015experimental}. The dataset comprises 1,797 drug molecules with continuous labels. 
\item VD. The volume of distribution at steady state (VDss) quantifies the extent to which a drug is distributed in body tissues relative to its concentration in the blood. A higher VDss indicates a greater distribution in the tissues and is typically associated with drugs that have high lipid solubility and low plasma protein binding rates~\cite{lombardo2016silico}. 
The dataset contains 1,130 drug molecules with continuous ground truth. 
\end{itemize}
\item {\bf Metabolism (M):} 
\begin{itemize}[leftmargin=*]
\item CYP2D6-I. The CYP P450 genes play a crucial role in the metabolism of diverse molecules and chemicals inside cells. One particular gene, CYP2D6, is predominantly expressed in the liver and exhibits high expression in specific regions of the central nervous system, such as the substantia nigra. Its function involves both the formation and breakdown of various substances~\cite{veith2009comprehensive}. The dataset contains 13,130 drug molecules with binary ground truth indicating their CYP2D6 inhibition~\cite{veith2009comprehensive}. 
\item CYP3A4-I. CYP3A4 is a vital enzyme present primarily in the liver and intestine. Its primary function is to catalyze the oxidation of small foreign organic molecules (xenobiotics), including drugs and toxins, enabling their elimination from the body~\cite{veith2009comprehensive}. 
The dataset contains 12,328 drug molecules with binary labels indicating the CYP3A4 inhibition~\cite{veith2009comprehensive}. 
\item CYP2C9-I. CYP1A2 is predominantly located in the endoplasmic reticulum and its expression can be induced by certain polycyclic aromatic hydrocarbons (PAHs), some of which are present in cigarette smoke. This enzyme plays a role in metabolizing certain PAHs into carcinogenic intermediates. Additionally, CYP1A2 is involved in the metabolism of other xenobiotic substances such as caffeine, aflatoxin B1, and acetaminophen~\cite{veith2009comprehensive}. The dataset contains 12,092 drug molecules with binary labels about CYP1A2 inhibition~\cite{veith2009comprehensive}. 
\item CYP2D6-S. CYP2D6 is predominantly expressed in the liver and exhibits significant expression in various regions of the central nervous system, including the substantia nigra. Substrates are drugs that are metabolized by the enzyme~\cite{carbon2011selecting}. The dataset contains 664 drug molecules with binary labels (whether it is a substrate to the enzyme). 
\item CYP3A4-S. CYP3A4, a crucial enzyme in the body, predominantly resides in the liver and intestine. Its primary function is to metabolize small exogenous organic molecules (xenobiotics), including drugs and toxins, through oxidation, facilitating their elimination from the body~\cite{carbon2011selecting}. The dataset contains 667 drug molecules with binary labels (whether it is a substrate to the enzyme). 
\item CYP2C9-S. CYP P450 2C9 is extensively involved in the oxidative metabolism of various xenobiotic and endogenous substances. It serves as a key enzyme responsible for the biotransformation of drug substrates~\cite{carbon2011selecting}. The dataset contains 666 drug molecules with binary labels (whether it is a substrate to the enzyme). 
\end{itemize}
\item {\bf Excretion (E):} 
\begin{itemize}[leftmargin=*]
\item Half-Life. The half-life of a drug refers to the period of time required for the concentration of the drug in the body to decrease by half. It provides a measure of the duration of a drug's effects. 
The dataset contains 667 drug molecules with continuous labels (i.e., half-life duration)~\cite{obach2008trend}. 
\item CL-Micro. Drug clearance is the term used to describe the volume of plasma from which a drug is completely removed over a specific period of time. It quantifies the rate at which the active drug is eliminated from the body. The dataset provided is a curated collection from the ChEMBL database, consisting of experimental data on intrinsic clearance. The dataset was deposited by AstraZeneca and includes clearance measurements obtained from two types of experiments: hepatocytes (CL-Hepa) and microsomes (CL-Micro)~\cite{wenlock2015experimental}. CL-Micro contains 1,102 drug molecules with continuous labels (removing time). 
\item CL-Hepa. CL-Hepa contains 1,020 drug molecules with continuous labels (removing time)~\cite{wenlock2015experimental}. 
\end{itemize}
\item {\bf Toxicity (T):} 
\begin{itemize}[leftmargin=*]
\item hERG. The human ether-à-go-go related gene (hERG) plays a critical role in regulating the rhythm of the heart. Consequently, the inhibition of hERG by a drug can result in significant adverse effects. Therefore, it is essential to accurately predict the potential hERG liability during the early stages of drug design~\cite{wang2016admet}. The dataset consists of 648 drug molecules with binary labels (inhibition of hERG)~\cite{wang2016admet}. 
\item AMES. Mutagenicity refers to a drug's capacity to induce genetic changes. Drugs that can cause DNA damage have the potential to lead to cell death or other serious adverse effects. The Ames test, developed by Professor Ames, is the most commonly used assay for assessing the mutagenicity of compounds~\cite{xu2012silico}. This short-term bacterial reverse mutation assay is capable of detecting a wide range of compounds that can cause genetic damage and frameshift mutations. The dataset consists of 7,255 drug molecules with binary labels (whether the drug is mutagenic (1) or not)~\cite{xu2012silico}. 
\item DILI. Drug-induced liver injury (DILI) is a life-threatening liver disease resulting from the use of certain drugs, and it has been the leading cause of drug marketing withdrawals due to safety concerns for the past five decades (e.g., iproniazid, ticrynafen, benoxaprofen). The dataset used in this study is compiled from the U.S. FDA's National Center for Toxicological Research, providing valuable information on drug-induced liver injury cases~\cite{xu2015deep}. 
The dataset consists of 475 drug molecules with binary labels (whether they can cause liver injury (1) or not (0))~\cite{xu2015deep}. 
\item LD50. The lethal dose 50 (LD) for acute toxicity quantifies the minimum dose at which lethal adverse effects are observed. A higher dose corresponds to a greater lethality of the drug. The dataset consists of 7,385 drug molecules with continuous labels (minimum dose for adverse effects)~\cite{zhu2009quantitative}. 
\end{itemize}
\end{itemize}

\begin{table*}[t!]
\centering
\caption{
Data statistics of small-molecule ADMET property datasets. 
ADMET refers to absorption, distribution, metabolism, excretion, and toxicity, respectively. 
The results are determined by wet-lab experiments. 
Due to the high expenses of wet-lab experiments and the data privacy issue, ADMET property datasets have small sizes. The label column indicates the data type for the corresponding property. 
}
\vspace{2mm}
\resizebox{0.50\textwidth}{!}{  
\begin{tabular}{llcccccc}
\toprule[1pt]
Category & Dataset & \# of data  & label \\ \hline 
\multirow{6}{*}{Absorption} & {Caco2\_Wang}  & 906 & continuous \\
 & {HIA\_Hou} & 578  & binary \\
 & {Pgp\_Broccatelli} & 1,212  & binary \\
 & {Bioavailability\_Ma} &  640  & binary \\
 & {Lipophilicity\_AstraZeneca} &  4,200  & continuous \\
 & {Solubility\_AqSolDB} & 9,982  & continuous \\ 
\midrule 
\multirow{3}{*}{Distribution} & {BBB\_Martins} & 1,975  & binary \\
  &  {PPBR\_AZ} &  1,797  & continuous \\
  &  {VDss\_Lombardo} & 1,130  & continuous \\
\midrule 
\multirow{8}{*}{Metabolism} &  {CYP2C19\_Veith}  & 12,092  & binary \\
  &  {CYP2D6\_Veith} & 13,130  & binary \\
  &  {CYP3A4\_Veith}  & 12,328  & binary \\
  &  {CYP1A2\_Veith} &  12,579  & binary \\
  &  {CYP2C9\_Veith}  & 12,092  & binary \\
  &  {CYP2C9\_Substrate}  & 666  & binary \\
  &  {CYP2D6\_Substrate}  & 664  & binary \\
  &  {CYP3A4\_Substrate}  & 667  & binary \\ 
\midrule 
\multirow{3}{*}{Excretion}  &  {Half\_Life\_Obach}  & 667  & continuous \\
  &  {Clearance\_Hepatocyte\_AZ}  & 1,020  & continuous \\ 
  &  {Clearance\_Microsome\_AZ}  & 1,102  & continuous \\ 
\midrule 
\multirow{8}{*}{Toxicity}   & {LD50\_Zhu} & 7,385  & continuous \\
   & {hERG} &  648  & binary \\
   & {AMES} &  7,255  & binary \\
   & {DILI} &  475  & binary \\
   & {Skin\_Reaction} & 404  & binary \\
   & {Carcinogens\_Lagunin} & 278  & binary \\
   & {Tox21} & 7,831  & binary \\
   & {ClinTox} & 1,484  & binary \\
\bottomrule[1pt]
\end{tabular}
}
\label{table:admet_dataset}
\end{table*}

ADMET prediction requires models to generalize to domains unseen during training, i.e., molecules with a new scaffold structure that are structurally different from drugs used for training. To meet this need, we adopt the scaffold split to mimic the realistic drug discovery scenario. The main idea of scaffold split is to split training/validation/test dataset based on scaffolds of the chemical compounds to make sure that the test dataset covers a diverse set of unseen scaffolds to measure its generalization ability to unseen drugs. For all the datasets, the training/validation/testing ratio is set to 7/1/2, where we conduct 5 independent runs for each method on each task. In each run, the training and validation sets are randomly shuffled with various random seeds.

\section{Experiment}
\label{sec:experiment}

In this section, we elaborate on the empirical studies, including baseline methods, evaluation metrics, experimental results, and their analysis.

\subsection{Baseline Methods}
We include the following baseline models for small-molecule pharmaceutical property prediction. 
\begin{enumerate}[leftmargin=*]
\item \textbf{Morgan+MLP}. Morgan molecular fingerprint is a fixed-dimensional binary vector (1024 bit here). It is followed by multiple layer perceptron (MLP) to carry out either classification or regression tasks. 
MLP has three hidden layers, and the hidden sizes are 1024, 512, and 128, respectively. 
The model has 1477K learnable parameters. 
\item \textbf{SMILES+CNN}. It uses SMILES string as the molecular representation and the input feature, which is followed by a one-dimensional convolutional neural network (1D-CNN). 1D-CNN has three layers; the number of filters for the three layers is 32, 64, and 96, respectively. The kernel sizes are 4, 6, and 8, respectively. After the convolutional layer, the hidden state is fed into a two-layer MLP whose latent dimensions are 32. 
The model has 227K learnable parameters. 
\item \textbf{GCN}. Graph convolutional network (GCN)~\cite{kipf2016semi} represents drug molecules in a molecular graph, where each atom corresponds to a node and each chemical bond corresponds to an edge. 
GCN has five layers, and the dimension of node embedding is set to 100. After GCN, all the node embeddings are aggregated with a summation function to get molecular-graph-level embedding, followed by a one-layer MLP to get the final prediction. 
The model has 192K learnable parameters. 
\item \textbf{NeuralFP}. NeuralFP uses Graph convolutional network (GCN)~\cite{kipf2016semi} to learn a neural network-based molecular embedding (also known as molecular \textit{neural fingerprint}, or NeuralFP) from a large amount of molecule data without labels~\cite{neuralfp}. The neural fingerprint is essentially a real-valued vector, also known as embedding. Then, the neural fingerprint is fixed and fed into a three-layer MLP to make the prediction. 
The hidden state dimensions are 200, 100, and 50. 
The model has 480K learnable parameters. 
\end{enumerate}

\subsection{Evaluation Metrics}

Drug pharmaceutical property prediction can be categorized into two machine learning tasks (classification and regression) based on the groundtruth. 
For classification tasks (mostly binary classification), we select one of the following two evaluation metrics based on the dataset: 
\begin{itemize}[leftmargin=*] 
\item \textbf{PR-AUC} (Precision-Recall Area Under Curve) summarizes the trade-off between the true positive rate and the positive predictive value for a predictive model using different probability thresholds. It is used for imbalanced data, e.g., the number of positives is smaller than the negatives. 
\item \textbf{ROC-AUC} (Area Under the Receiver Operating Characteristic Curve) summarizes the trade-off between the true positive rate and the false positive rate for a predictive model using different probability thresholds. It is typically used for balanced data, where the number of positive and negative samples is close. 
\end{itemize} 
For both PR-AUC and ROC-AUC, higher values are more desirable. 
On the other hand, for regression tasks, we select one of the following two evaluation metrics based on the dataset: 
\begin{itemize}[leftmargin=*]
\item \textbf{Mean Absolute Error (MAE)} measures the absolute value of the difference between the predicted value and the actual value. A lower MAE value indicates better performance. 
\item \textbf{Spearman's rank correlation coefficient (Spearman)} is the Pearson correlation coefficient between the rank variables. Higher values indicate better performance. It is used when a trend (ranking) is more important than the absolute error. 
\end{itemize}

\begin{table*}[t!]
\centering
\caption{
Performance of various machine learning methods on drug \textbf{absorption} property prediction tasks. 
The absorption property describes how drugs are absorbed into the human body to reach the site of action~\cite{benet1996pharmacokinetics}. 
Average and standard deviation across five runs are reported. 
The arrow $\downarrow$ in the bracket indicates a lower score is better, while $\uparrow$ indicates the opposite. 
On each task, the best method is bolded, and the second best is underlined. 
}
\resizebox{\textwidth}{!}{  
\begin{tabular}{cccccccc}
\toprule[1pt]
Dataset & {Caco2} & {HIA} & {Pgp} & {Bioav} & {Lipo} &  {AqSol} \\ 
 \midrule  
Size & 906 & 578 & 1,212 & 640 & 4,200 & 9,982 \\ \midrule  
Metric & MAE ($\downarrow$) & ROC-AUC ($\uparrow$) & ROC-AUC ($\uparrow$) & ROC-AUC ($\uparrow$) & MAE ($\downarrow$) & MAE ($\downarrow$) \\ \hline 
Morgan+MLP & 0.908\std{0.060} & 0.807\std{0.072} & 0.880\std{0.006} & 0.581\std{0.086} & {0.701\std{0.009}} & 1.203\std{0.019}  \\ 
SMILES+CNN & \underline{0.446\std{0.036}} & 0.869\std{0.026} & 0.908\std{0.012} & 0.613\std{0.013} &  0.743\std{0.020} & 1.023\std{0.023}   \\ 
GCN & 0.599\std{0.104} & 0.936\std{0.024} & 0.895\std{0.021} & 0.566\std{0.115} & {0.541\std{0.011}} & \underline{0.907\std{0.020}}   \\ 
NeuralFP & 0.530\std{0.102} & \textbf{0.943\std{0.014}} & \underline{0.902\std{0.020}} & \underline{0.632\std{0.036}} & \underline{0.563\std{0.023}} & \textbf{0.947\std{0.016}}  \\ 
\mname~& \textbf{0.438\std{0.030}} & \underline{0.937\std{0.011}} & \textbf{0.930\std{0.017}} & \textbf{0.673\std{0.025}} & \textbf{0.583\std{0.020}} & 0.819\std{0.020} \\ 
\bottomrule[1pt]
\end{tabular}
}
\label{tab:admet_absorption}
\end{table*}

\begin{table*}[t!]
\centering
\caption{
Performance of various machine learning methods on drug \textbf{distribution} property prediction tasks. 
The distribution property is important as it affects the drug's concentration at the target site, efficacy, and potential side effects. Factors influencing drug distribution include lipophilicity (ability to dissolve in lipids), molecular size, binding to plasma proteins, tissue permeability, and the presence of efflux transporters~\cite{benet1996pharmacokinetics}. 
Average and standard deviation across five runs are reported. 
The arrow $\downarrow$ in the bracket indicates a lower score is better, while $\uparrow$ indicates the opposite. 
On each task, the best method is bolded, and the second best is underlined. 
}
\vspace{1mm} 
\resizebox{0.567\textwidth}{!}{  
\begin{tabular}{cccc}
\toprule[1pt]
Dataset & {BBB} & {PPBR} & {VD} \\ 
\midrule 
Size & 1,975 & 1,797 & 1,130 \\ 
\midrule 
Metric  & ROC-AUC ($\uparrow$) & MAE ($\downarrow$) & Spearman ($\uparrow$) \\ \hline 
Morgan+MLP & 0.823\std{0.015} & 12.848\std{0.362} & \underline{0.493\std{0.011}} \\ 
SMILES+CNN & 0.781\std{0.030} & 11.106\std{0.358} & 0.226\std{0.114} \\ 
GCN & \underline{0.842\std{0.016}} & 10.194\std{0.373} & 0.457\std{0.050} \\ 
NeuralFP & 0.836\std{0.009} & \bf 9.292\std{0.384} & 0.258\std{0.162} \\ 
\mname~& \textbf{0.852\std{0.018}} & \underline{9.371\std{0.311}} & \textbf{0.471\std{0.099}} \\  
\bottomrule[1pt]
\end{tabular}
}
\label{tab:admet_distribution}
\end{table*}

\begin{table*}[t!]
\centering
\caption{
Performance of various machine learning methods on drug \textbf{metabolism} property prediction tasks. 
The metabolism property refers to the process by which a drug undergoes chemical transformations in the body, primarily in the liver, to be converted into metabolites~\cite{benet1996pharmacokinetics}. 
Average and standard deviation across five runs are reported. 
The arrow $\downarrow$ in the bracket indicates a lower score is better, while $\uparrow$ indicates the opposite. 
On each task, the best method is bolded, and the second best is underlined. 
}
\vspace{1mm}
\resizebox{\textwidth}{!}{  
\begin{tabular}{ccccccc}
\toprule[1pt]
Dataset & {CYP2D6-I} & {CYP3A4-I} & {CYP2C9-I} & {CYP2D6-S} & {CYP3A4-S} & {CYP2C9-S} \\ 
 \midrule  
Size & 13,130 & 12,328 & 12,092 & 664 & 667 & 666 \\ 
\midrule   
Metric  & PR-AUC ($\uparrow$) & PR-AUC ($\uparrow$) & PR-AUC ($\uparrow$) & PR-AUC ($\uparrow$) & ROC-AUC ($\uparrow$) & PR-AUC ($\uparrow$) \\ \hline 
Morgan+MLP & 0.587\std{0.011} & {0.827\std{0.009}} & 0.715\std{0.004} & \underline{0.671\std{0.066}} & 0.633\std{0.013} & \bf 0.380\std{0.015} \\ 
SMILES+CNN & 0.544\std{0.053} & 0.821\std{0.003} & 0.713\std{0.006} & 0.485\std{0.037} & \underline{0.662\std{0.031}} & \underline{0.367\std{0.059}} \\ 
GCN & 0.616\std{0.020} & 0.840\std{0.010} & 0.735\std{0.004} & 0.617\std{0.039} & 0.590\std{0.023} & 0.344\std{0.051} \\ 
NeuralFP & \underline{0.627\std{0.009}} & \underline{0.849\std{0.004}} & \underline{0.739\std{0.010}} & 0.572\std{0.062} & 0.578\std{0.020} & 0.359\std{0.059} \\ 
\mname~& \textbf{0.747\std{0.013}} & \textbf{0.893\std{0.012}} & \textbf{0.845\std{0.011}} & \textbf{0.748\std{0.012}} & \textbf{0.664\std{0.027}} & 0.365\std{0.021} \\ 
\bottomrule[1pt]
\end{tabular}
}
\label{tab:admet_metabolism}
\end{table*}

\begin{table*}[t!]
\centering
\caption{
Performance of various machine learning methods on drug \textbf{excretion} property prediction tasks. 
The excretion property refers to the process by which drugs and their metabolites are eliminated from the body~\cite{benet1996pharmacokinetics}. Average and standard deviation across five runs are reported. The arrow $\downarrow$ in the bracket indicates a lower score is better, while $\uparrow$ indicates the opposite. On each task, the best method is bolded, and the second best is underlined. }
\vspace{1mm}
\resizebox{0.56\textwidth}{!}{  
\begin{tabular}{cccc}
\toprule[1pt]
Dataset & {Half-Life} & {CL-Micro} & {CL-Hepa} \\ \midrule 
Size & 667 & 1,102 & 1,020 \\
 \midrule 
Metric  & Spearman ($\uparrow$) & Spearman ($\uparrow$) & Spearman ($\uparrow$) \\ \hline 
Morgan+MLP & \bf 0.329\std{0.083} & 0.492\std{0.020} & 0.272\std{0.068} \\ 
SMILES+CNN & 0.038\std{0.138} & 0.252\std{0.116} & 0.235\std{0.021} \\ 
GCN & \underline{0.239\std{0.100}} & \bf 0.532\std{0.033} & 0.366\std{0.063} \\ 
NeuralFP & 0.177\std{0.165} & \underline{0.529\std{0.015}} & \underline{0.401\std{0.037}} \\ 
\mname~& 0.247\std{0.100} & 0.501\std{0.049} & \bf 0.423\std{0.029} \\ 
\bottomrule[1pt]
\end{tabular}
}
\label{tab:admet_excretion}
\end{table*}

\begin{table*}[t!]
\centering
\caption{Performance of various machine learning methods on drug \textbf{toxicity} property prediction tasks. 
The toxicity property refers to the potential adverse effects or harmful interactions that a drug or its metabolites may have on living organisms, including humans~\cite{benet1996pharmacokinetics}. 
Average and standard deviation across five runs are reported. 
The arrow $\downarrow$ in the bracket indicates a lower score is better, while $\uparrow$ indicates the opposite. 
The best method is bolded on each task, and the second best is underlined. }
\resizebox{0.7\textwidth}{!}{  
\begin{tabular}{ccccc}
\toprule[1pt]
Dataset & {hERG} & {AMES} & {DILI} & {LD50} \\ \hline 
Size & 648 & 7,255 & 475 & 7,385 \\  
\midrule  
Metric  & ROC-AUC ($\uparrow$) & ROC-AUC ($\uparrow$) & ROC-AUC ($\uparrow$) & MAE ($\downarrow$) \\ \hline 
Morgan+MLP & \underline{0.736\std{0.023}} & 0.794\std{0.008} & 0.832\std{0.021} & 0.649\std{0.019} \\ 
SMILES+CNN & \textbf{0.754\std{0.037}} & 0.776\std{0.015} & 0.792\std{0.016} & \underline{0.675\std{0.011}} \\ 
GCN & 0.738\std{0.038} & \underline{0.818\std{0.010}} & \underline{0.859\std{0.033}} & 0.649\std{0.026}  \\ 
NeuralFP & 0.722\std{0.034} & \textbf{0.823\std{0.006}} & 0.851\std{0.026} & 0.667\std{0.020} \\ 
\mname~& 0.708\std{0.045} & {0.801\std{0.030}} & \textbf{0.928\std{0.022}} & \textbf{0.678\std{0.012}} \\ 
\bottomrule[1pt]
\end{tabular}
}
\label{tab:admet_toxicity}
\end{table*}

\subsection{Implementation Details}
All the experiments are conducted on an NVIDIA GeForce RTX 3090 GPU. \mname is implemented in Python 3.8 and PyTorch 1.9.0. The Adam~\cite{kingma2014adam} is used as a numerical optimizer with an initial learning rate of 1e-3. In the pretraining phase, the maximal epoch number is set to 100. In the fine-tuning process, the maximal epoch number is set to 50. We use an early stop strategy to save training time and computational resources and avoid overfitting. The \mname~architecture consists of 8 layers, the hidden state dimension is set to 300, while the number of attention heads is set to 8.

\subsection{Results \& analysis.}

The results for absorption, distribution, metabolism, excretion, and toxicity property prediction are reported in Table~\ref{tab:admet_absorption},~\ref{tab:admet_distribution},~\ref{tab:admet_metabolism},~\ref{tab:admet_excretion} and~\ref{tab:admet_toxicity}, respectively. 
For some of the baseline methods, we reuse the results already reported in Therapeutics Data Commons' Benchmark~\cite{huang2021therapeutics,huang2022artificial}. We follow the same setup with them for fair comparison. 
By carefully comparing all the results, we draw a couple of conclusions as follows, 

\begin{itemize}
\item First, the proposed \mname~model exhibits brilliant performance in all 22 ADMET tasks. Concretely, compared with 4 cutting-edge machine learning models, it achieves the highest score in 14 tasks and top-2 performance in 17 tasks among all the 22 tasks. 
\item Second, self-supervised learning-based pretraining strategies prove to be highly effective. Specifically, models like the proposed \mname~and NeuralFP~\cite{neuralfp} demonstrate exceptional performance by leveraging self-supervised learning to extract valuable insights from unlabeled data. These approaches highlight the potential of self-supervised learning as a promising direction for future research, indicating its significant impact on enhancing model performance in molecular ADMET property prediction. 
\item Third, no single method dominates all tasks, as performance varies depending on the feature types and the specific tasks at hand. This variation arises from the different kinds of information that various molecular representations and machine learning models capture. For example, GNN models like GCN and NeuralFP focus on local substructures within molecular graphs, while the CNN model captures broader biochemical features from SMILES strings. Consequently, integrating these diverse feature representations has the potential to further enhance model performance.
\end{itemize}

\section{Conclusion}
In this paper, we introduced \mname, a novel two-stage model designed for drug ADMET property prediction by leveraging both unlabeled and labeled data. Through a combination of self-supervised pretraining and fine-tuning, \mname~effectively captures the underlying chemical structures and relationships inherent in molecular data. Our extensive experiments demonstrated that \mname~outperforms several state-of-the-art models across a range of ADMET tasks, highlighting the efficacy of self-supervised learning in molecular property prediction. By reducing the reliance on large, labeled datasets, this approach not only enhances prediction accuracy but also offers a promising direction for future research in drug discovery, potentially accelerating the identification and development of safe and effective drug candidates. The success of \mname~underscores the importance of advanced machine learning techniques in addressing the complex challenges of drug discovery and development.

Future work can be conducted in following three aspects: (1) During early-stage clinical trials, precise ADMET profiling helps researchers understand how a drug is absorbed, distributed within the body, metabolized by enzymes, excreted, and whether it poses any toxic risks. This detailed information allows for the identification of potential safety concerns before large-scale trials begin, helping to prevent costly failures at later stages~\cite{yue2024ct,lu2024uncertainty,chen2024uncertainty}; (2) integration of ADMET data with multi-omics: researchers can gain deeper insights into how genetic, transcriptomic, and metabolic variations influence drug behavior and response in different individuals or populations~\cite{lu2019integrated,lu2021cot,lu2022cot,chen2021data}. This combination enables the identification of biomarkers for predicting drug efficacy and toxicity, supports the development of more effective and personalized therapeutics, and helps to minimize adverse drug reactions~\cite{fu2024ddn3,zhang2021ddn2}.

{\small
\bibliographystyle{plain}
\bibliography{main.bib}
}

\end{document}

%% file: setup/package.tex
\usepackage{graphicx}
\usepackage{subcaption}
\usepackage{float}
\usepackage[justification=raggedright]{caption}	
\usepackage{lscape}                                         

\usepackage[lined,ruled,linesnumbered]{algorithm2e}

\usepackage{booktabs}                   
\usepackage{multirow}

\usepackage{paralist}
\usepackage{enumitem}

\usepackage{bm}                          
\usepackage{epsfig}                      
\usepackage{graphicx}                  
\usepackage{times}
\usepackage{mathptmx}
\usepackage{mathtools}
\usepackage{amssymb,amsmath}   

\usepackage{units}
\usepackage{color}

\usepackage{comment}

\usepackage{url}  
\usepackage[pagebackref=true,breaklinks=true,letterpaper=true,colorlinks,bookmarks=false]{hyperref}

\usepackage{xspace}
\usepackage[table]{xcolor}
\usepackage{setspace}


%% file: setup/macros.tex
\newlength\paramargin
\newlength\figmargin
\newlength\secmargin

\long\def\ignorethis#1{}

%% file: setup/symbols.tex
\def\xi{\mathbf{x}_i}

%% file: setup/graphicspath.tex
\graphicspath{{figure}, {example}}